# Real-time AdaBoost cascade face tracker based on likelihood map and optical flow

Andreas Ranftl[1], Fernando Alonso-Fernandez[1] ✉, Stefan Karlsson[1], Josef Bigun[1]
[1]School of Information Technology, Halmstad University, Box 823, Halmstad SE 301-18, Sweden
✉ E-mail: feralo@hh.se

**Abstract:** The authors present a novel face tracking approach where optical flow information is incorporated into a modified version of the Viola–Jones detection algorithm. In the original algorithm, detection is static, as information from previous frames is not considered; in addition, candidate windows have to pass all stages of the classification cascade, otherwise they are discarded as containing no face. In contrast, the proposed tracker preserves information about the number of classification stages passed by each window. Such information is used to build a likelihood map, which represents the probability of having a face located at that position. Tracking capabilities are provided by extrapolating the position of the likelihood map to the next frame by optical flow computation. The proposed algorithm works in real time on a standard laptop. The system is verified on the Boston Head Tracking Database, showing that the proposed algorithm outperforms the standard Viola–Jones detector in terms of detection rate and stability of the output bounding box, as well as including the capability to deal with occlusions. The authors also evaluate two recently published face detectors based on convolutional networks and deformable part models with their algorithm showing a comparable accuracy at a fraction of the computation time.

## 1 Introduction

Face detection and tracking are two important tasks in many fields including biometrics, human–computer interaction, video surveillance or forensics. They are fundamental techniques for a computer to interact with humans in a natural way, and they have a major influence on the performance of all facial analysis algorithms including face alignment, two-dimensional (2D) and 3D face modelling, face recognition, head tracking or gender/age recognition [1]. Face analysis is being driven by the trend and popularity of social networking sites (with companies such as Apple, Google, Microsoft or Facebook including face recognition in their products), the prevalence of mobile smart-phone applications and low-cost sensors, and the successful implementation in visa application and in criminal and military investigation [2]. Face tracking describes the process of following a face through every single frame of a video, for which it must first be detected. Some methods also employ face recognition to ensure that the person in the frame is the same person as in the previous frame [3], but this aspect will not be considered here.

Viola and Jones introduced a real-time face detector in 2001 [4], which has evolved as an appreciated tool of computer vision and baseline algorithm [5]. Face detection is performed by applying a cascade of classifiers on several windows within the image. The windows vary in location and size in order to determine scale and position of the face. The Viola–Jones method does not incorporate temporal constraints, nor combines evidence from previous frames to aid in the inference. In short, it is a static algorithm which restarts face detection in each new frame. Apart from lacking temporal consistency, it does not save information about near-positives either (candidate windows that pass a significant number of stages of the classification cascade). Not only positions of detected faces may behave erratically due to restarting detection in every frame, but if a face becomes temporarily distorted or occluded, so that the very last part of the cascade fails, the detection fails abruptly.

This paper investigates a way to extend the Viola–Jones method to incorporate information about near-positives, as well as to include tracking capabilities. We extend our previous study presented in a conference paper [6]. The developed tracker takes advantage of optical flow calculations between frames. The results of the optical flow computations are used to interpolate a likelihood map with respect to time. The likelihood map describes the probability for a pixel being part of a face. It is built-up by summing up the number of classification stages that a window has passed during evaluation by the Viola–Jones algorithm. This way, information about near-positives is preserved, even if they do not pass all stages of the cascade of classifiers. Our face tracker calls this modified version of the Viola–Jones algorithm every $n$ frames, and in the frames in between, face detection and tracking is done by interpolating the likelihood map from frame to frame with optical flow calculations. The proposed face tracking system shows a higher detection rate w.r.t. Viola–Jones, while keeping a good accuracy. More importantly, it enables face detection and tracking under partial or complete occlusion, and our system has less false negatives (i.e. face losses) due to the use of the proposed likelihood map. We also achieve less erratic movements in the bounding box surrounding the detected faces. Our system also compares well in terms of accuracy to two recently published algorithms based on multi-task cascaded convolutional networks (MTCNNs) [7] and deformable part models [8, 9]. With an accuracy comparable to these newer approaches, our system has a detection speed one order of magnitude smaller.

The remaining of this section presents a literature review on the topics of face detection and tracking, as well as the contributions of this paper. Section 2 describes the proposed face detection and tracking system. The experimental framework, including baseline detection methods, database, performance metrics and results, is given in Section 3, followed by conclusions in Section 4. A supporting video can be found at https://youtu.be/U7PiQu11Auo.

### 1.1 Literature review

*1.1.1 Face detection:* Several algorithms have been developed for face detection [5, 10], mostly based on scanning the image with sub-windows in a raster-like fashion, and classifying the pattern in the sub-window as face/non-face. Scale change is overcome with image pyramids, a time-consuming approach. The face/non-face classifier is learned from face/non-face training sub-windows using statistical learning methods. The rationale behind this is that pixels on a face are highly correlated, whereas those in a non-face sub-window present much less regularity [5]. Variations caused by lightning or illumination, face expression, pose or appearance make necessary the use of non-linear classifiers. Examples are neural networks [11, 12] or support vector machines [13]. However, the breakthrough in face detection happened with Viola–Jones detector [14]. This algorithm has evolved as a de-facto standard [5], with an implementation included in OpenCV.

The Viola–Jones detector uses a set of Haar features for image scanning, which resemble features such as centre-surround and directional responses. A variant of the AdaBoost learning algorithm is used to select strong features and to combine weak classifiers in cascade into strong ones, in a way that background regions are quickly discarded while spending more time (i.e. classifiers) on candidate, object-like windows. If a window passes all the classification stages, then it is considered to contain a face. Accuracy and speed are balanced with the number of classifiers and its individual performance, so a non-face candidate window is rejected in the early stages of the classification cascade. Since it is expected that a face is detected by several overlapped windows that differ slightly in position and/or scale, the number of false positives is further reduced by imposing a minimum number of windows that must overlap in order to have a face. Position of the detected face is thus refined by combining all the overlapped windows involved. The Viola–Jones algorithm is able to work in real time, and it has proven to work very well with near-frontal faces, but its performance is significantly decreased for faces at arbitrary poses and other degradations [15]. Multi-view face detection is usually dealt by training separate classifiers for each face view, but if pose estimation fails, e.g. a face is misclassified as frontal, it may never be detected by the frontal face cascade, not to mention the increased computational complexity [5]. Subsequent improvements of the Viola–Jones algorithm have included the use of more complex features to obtain a more accurate classification, such as diagonal Haar-like feature sets [16] or multi-block local binary patterns (MB-LBP) [17]. The latter, which is regarded as faster and more discriminative than Haar-like features by its authors, is the implementation that we will employ in this paper.

More recently, the cascade framework (i.e. quick rejection of background regions) has been applied to features learned by convolutional neural networks (CNN) [18], and the amount of research works on face detection making use of CNNs is exploding, e.g. [7, 19–21], inspired by the remarkable recent success of CNNs in many computer vision tasks. A drawback of these approaches, and of many approaches for unconstrained face detection, is that they usually need a considerable amount of annotated training data, apart from being computationally expensive [7]. Unconstrained face detection has gained attention recently also due to benchmarks for the development of unconstrained face detection algorithms, such as the face detection data set and benchmark (FDDB) [22], or the annotated facial landmarks in the wild (AFLW) [23]. Proposed approaches for unconstrained face detection include, for example, SURF features in an AdaBoost cascade [24], probabilistic elastic parts [25], deformable parts models (DPM) [9] or normalised pixel differences in a deep quadratic tree [26]. However, accuracy of unconstrained face detection is not yet satisfactory, especially when the detector is required to have low false alarms [26].

*1.1.2 Face tracking:* A number of algorithms to track detected faces in videos have been proposed in the literature. Face tracking algorithms can be classified in several ways [27], for example considering if the whole face is tracked as a single entity (head tracking), if individual face features are tracked instead (feature tracking), or whether the tracking is done in the 2D or in the 3D space (with the latter facilitating estimation of the 3D pose). To track facial features, it is useful to know the head pose in order to have a rough estimate of the location of the features, but on the other hand it is also possible to estimate head pose by using the detected/tracked facial features. It is usual then that detection and tracking of the whole face and individual features is done simultaneously in a two-level hierarchy, so one approach complements the other, as for example if active appearance models are used [28–30]. Among the main challenges that a face tracking system has to solve, they include variations of pose and lightning, facial deformation, occlusions or real-time capabilities [31]. Most of the methods only address one or two of these challenges, and many of them need to be trained or initialised manually [32]. Several authors have proposed the use of optical flow estimation to aid in the track of faces between frames [32–37]. Other works, not discussed here, make use for example of mean shift [38–43], particle filters [44–47], Kalman filters [48–50], active contour models [34, 35, 51], wavelet transform [52] or Bayesian networks [53], or Markov random fields [54]. This list is not exhaustive but only gives works published after 2000. We refer the reader to [55, 56] for additional information. Recently, unconstrained correlation filters based method shows a promising performance on benchmark datasets [57, 58], with works on face tracking making use of this technique too [59–63].

The optical flow describes the displacement of features from one image to another image within an image sequence [64], giving the spatial offset. Two main types of optical flow methods exist: sparse and dense. Dense flow methods operate on all pixels of the image, whereas sparse methods operate only on a subset of them. Those using sparse data rely heavily upon existence of discriminatory and identifiable local features. An example is the algorithm of Shi and Tomasi [65], which selects points possessing significant image gradients in two orthogonal directions, which are considered good features for tracking. A neighbourhood containing two orthogonal directions occurring equally frequently is known as a perfectly balanced neighbourhood [66]. It is called so, because translation of the neighbourhood in an arbitrary direction will produce the same error (measured in L2 norm), which also happens to be the maximal error, the translational error signalling isotropically the inflicted displacement maximally. Such neighbourhoods are also known as lacking linear symmetry [66] or corner-like [67], and can be conveniently found via eigenvalues of the structure tensor in image neighbourhoods of any dimension, not just two, and even in non-Cartesian coordinates [64].

Other inherent problem of sparse methods is correspondence, i.e. which sparse point corresponds to which between two frames, along with the occlusion of points by moving objects. In the work of [33], face detection is done by applying 3D deformable face models. Optical flow information (computed from a sparse set of points detected via [65]) is combined with edge information using a Kalman filter, with the aim of minimising the noise which is present in the optical flow calculations. The algorithm, however, is not suitable for real-time processing. The method in [37] is based on the principle of tracking–learning–detection (TLD). Each frame is processed by a frontal face detector and a tracker, and their outputs are passed to an integrator which estimates the object location. According to the TLD principle, the face under tracking is simultaneously learned, so the detector is updated with information of the object being tracked. Displacement of sparse candidate patches for tracking is estimated with the Lucas–Kanade method [68]. The work in [32] tracks mouth and eyebrow patches after the head is detected, with the aim of capturing facial expressions. Other works have dealt with the issue of tracking the mouth for lip reading purposes by optical flow using active contours [34, 35] or shape-from-motion features [69]. Tracking the mouth of a talking person is useful for many applications such as identity recognition or speechless human–computer interaction. Lastly, the method in [36] tracks pupils and nostrils as well by optical flow, in addition to mouth corners.

*1.2 Contributions*

This paper extends our previous study [6] with additional experimental results. The face detection and tracking system is described more formally and in greater detail. The developed method utilises a modified version of the well-known Viola–Jones algorithm for face detection, which is complemented with the addition of tracking capabilities via optical flow computation. We also include two recently published algorithms based on MTCNNs [7] and deformable part models [8, 9] to our comparison framework.

Within the Viola–Jones method, classification is done with a cascade of 20 stages, and a candidate window has to pass all the stages to be considered a face window. It does not keep information about near-positives, and does not consider detection

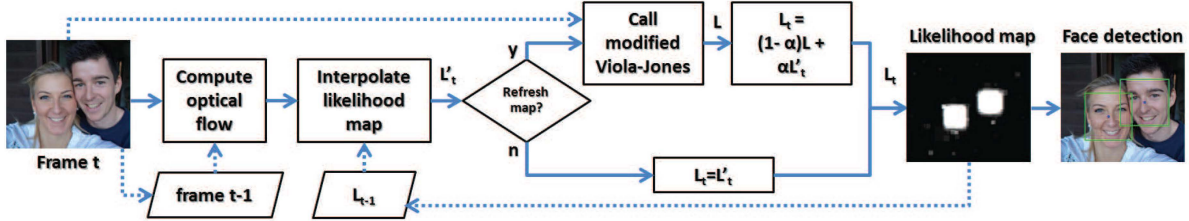

**Fig. 1** *Flow chart of the developed optical flow based face tracker. The likelihood map $L_t$ of the input frame and the detected faces are also shown. Grey intensity in the likelihood map indicates probability (re-scaled to 0-255 for visualisation purposes)*

information from previous frames. Instead, Viola–Jones restarts face detection in each new frame. In contrast, the developed approach calls Viola–Jones method only every $n > 1$ frames for refreshing the likelihood map. We modify the OpenCV implementation of the Viola–Jones face detector in the way that a likelihood map is built. The likelihood values within this map are dependent on the numbers of classification stages of the cascade that each detection window passes. In order to track the faces and preserve information from previous frames, position of the likelihood map in the next frame is interpolated via dense flow map. When Viola–Jones is triggered again, the likelihood map is refreshed by combination of the likelihood map interpolated from the previous frame, and the newly generated likelihood map from the modified Viola–Jones cascade. This way, information from previous features is not dismissed when Viola–Jones is called again.

Compared with the original Viola–Jones implementation, our approach enables faces to be detected even when they do not pass all of the stages of the cascade classifier. An image region gets also a high value in the likelihood map if the respective window passes a certain number of classification stages, and the likelihood map is populated in proportion to the number of stages passed by the window. This way, information about the reliability of detection of each window is preserved, in contrast to the *all-or-nothing* approach of Viola–Jones. Another advantage of the developed face tracker is that it can also track faces during partial occlusion and even under complete occlusion, due to interpolation of the likelihood map from frame to frame by optical flow. Our approach is also one order of magnitude faster than the two other algorithms evaluated in this paper, while keeping a similar detection accuracy, something that can be an advantage in environments with limited computing resources.

## 2 Optical flow enhanced AdaBoost cascade face detector and tracker

### 2.1 Overview of the proposed algorithm

A flowchart of the proposed system can be found in Fig. 1. In the proposed algorithm, the Viola–Jones algorithm is executed every $n$ frames (called *refresh* frame), but instead of using its binary output of detected/non-detected faces, we produce a likelihood map which computes for every pixel the probability of having a face located at that respective position. The mentioned probability is derived from the number of classifiers of the cascade that every image window has passed, and face position is then extracted from the likelihood map by thresholding. In addition, the optical flow between two consecutive frames is computed. In intermediate frames where Viola–Jones is not executed, position of the new likelihood map is interpolated according to optical flow computations, and it is refreshed by recursive filtering after $n$ frames, when Viola–Jones is re-executed. If no face can be detected in a refresh frame, the modified Viola–Jones is executed again in the next frame, with this procedure repeated until a frame with a face is obtained. In the meantime, the likelihood map is updated frame by frame using optical flow information.

The proposed setup allows to detect a face window as long as it passes a sufficient number of classifiers, and prevents that it is lost between frames by means of dense optical flow information even in case of occlusion. Additionally, size of the face window does not change erratically as in the original Viola–Jones algorithm, which handles face detection frame by frame separately, so that the face window may appear unstable even if the face does not move.

### 2.2 Likelihood map computation

When Viola–Jones is executed at time $t$, a refresh likelihood map $L$ is built by summing up at each pixel the number of stages of the Viola–Jones classification cascade passed by each detection window. In order to preserve temporal information from the previous frames, the likelihood map $L_t$ at time $t$ is then computed recursively as

$$L_t = (1 - \alpha) \times L + \alpha \times L'_t \qquad (1)$$

where $L'_t$ is the likelihood map $L_{t-1}$ at time $t-1$, interpolated using the optical flow between frames at $t-1$ and $t$. If Viola–Jones does not need to be called at time $t$, the likelihood map is just updated as $L_t = L'_t$.

Time savings can be achieved by computing $L$ using only windows that have passed a certain number $\tau$ of classifiers. To compute $L$, we set every pixel of the detection window to the corresponding number of classification stages that have been passed by that window. Fig. 1 (right) gives an example of a likelihood map from an image with two faces. All detection windows having passed a number $\tau$ of classifiers are summed up, but before a window is added, it is shrunk by a factor $s$. This is because Viola–Jones detection windows are usually bigger than the faces outlined by them, including pixels that do not contain a face. Since pixels not belonging to the face usually will not contain motion that is concurrent with the head movement, boundaries of the likelihood map would not be correctly interpolated with the optical flow if these pixels are kept. At the same time, this allows to separate windows from faces that appear adjacent to each other.

### 2.3 Likelihood map interpolation by optical flow

We use the Farnebäck dense flow method [70], which is suitable for real-time applications and exists as ready-to-use function in OpenCV. Dense flow methods compute displacement of all pixels of the image. We employ dense flow as we seek pixel-wise interpolation of the likelihood map in order to account for non-rigidity of the face. Sparse flow is faster to compute, but it would not be suitable for the proposed interpolation between frames, since we want to interpolate the set of adjacent pixels that represent a face in the likelihood map (Fig. 1, right). Geometric 2D interpolation of $L_{t-1}$ to $L'_t$ is done using the OpenCV function `remap()`.

### 2.4 Face detection from the likelihood map

To extract the face area from $L_t$, we first apply a threshold $c$ to binarise the likelihood map. This is to avoid false positives at smaller peaks. We then label individual adjacent areas as different faces, and their centres are used as the associated face centres. To compute the bounding box outlining the face, the individual areas are stretched by the inverse of the shrinking factor $s$ applied to the windows before they were summed up to build the likelihood map. This procedure outputs a rectangle instead of an square (as it is the case of Viola–Jones), allowing a more accurate estimation of the face region. Fig. 1 shows the likelihood map of an image containing two adjacent faces, as well as the detected faces using

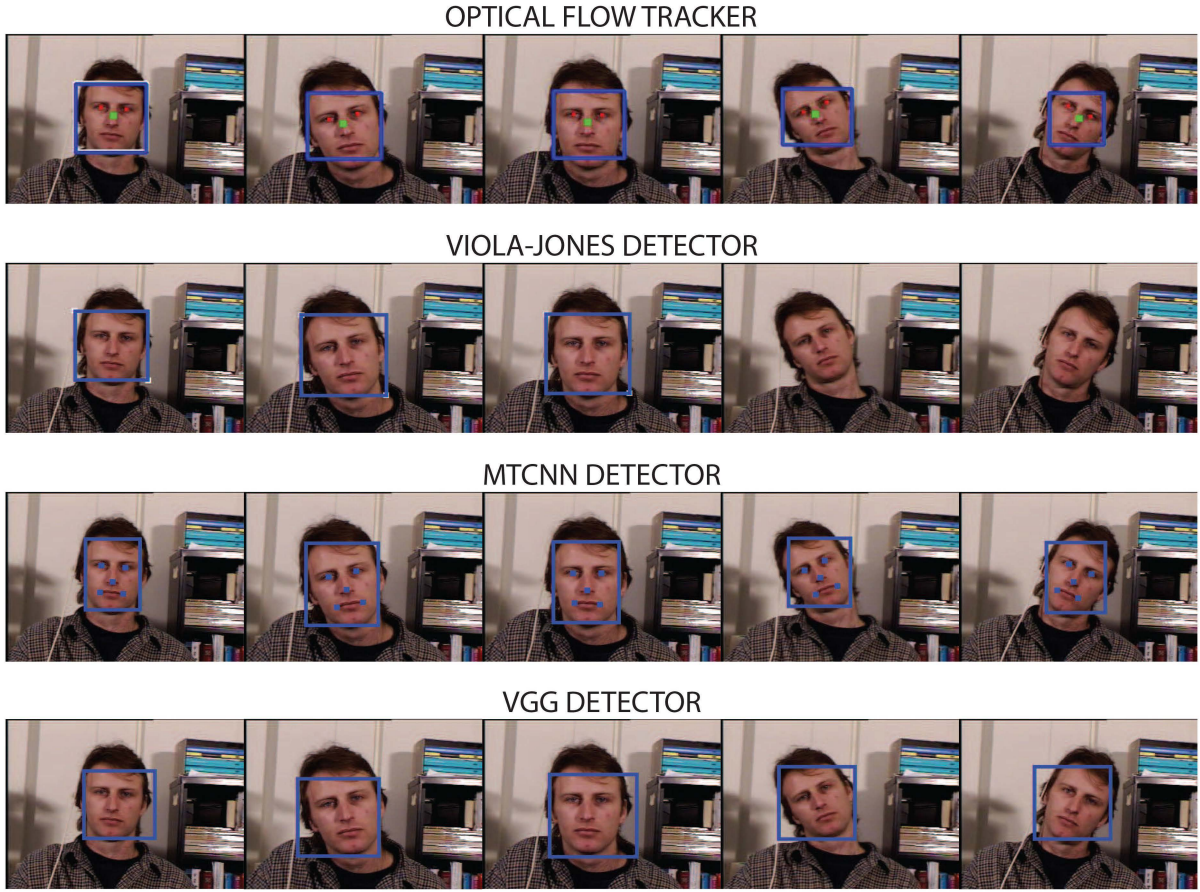

**Fig. 2** *Example frames (23, 92, 100, 150 and 188) of the video jam1.avi from the Boston Head Tracking Database. Red dots in the first row indicate ground-truth positions of the eyes, and green dots represent face centres computed from the ground truth. The blue rectangles and blue dots represent the output given by the respective detector (note that MTCNN also outputs facial landmarks, while the other detectors only output the face bounding box). In frame 23, a refresh is done by the Viola–Jones algorithm (denoted by the white bounding box in the top left image). In frames 150 and 188, the Viola–Jones algorithm resulted in a false negative (no face detected)*

the procedure indicated in this section. Areas containing the two faces show a high probability, which decreases towards the boundaries of the white regions.

## 3 Experimental framework

### 3.1 Baseline face detection methods

We also conduct face detection experiments using two recently published algorithms [7, 8]. The method in [7] is based on a MTCNN with three stages of deep convolutional networks (CNN) which predict face and landmark location in a coarse-to-fine manner. The complexity of the CNN increases from stage to stage, in a way that quick rejection of non-face windows is done at the beginning, then a more complex CNN is used in the last stage to refine the results. The training databases employed include WIDER FACE [71] and CelebA [19] databases, while evaluation results reported in [7] were obtained using FDDB [22], AFLW [23] and WIDER FACE databases. We use the code released by the authors https://kpzhang93.github.io/MTCNN_face_detection_alignment for our experiments, which is implemented using Caffe. Regarding the method of [8], it is based on the DPM algorithm of [9], which is included in the release of the VGG-Face CNN recognition descriptor http://www.robots.ox.ac.uk/~vgg/software/vgg_face described in [8], that we also use for our experiments here. The implementation used is run using MatConvNet. The method originally described in [9] was trained using images from AFLW [23] and Pascal Faces [72] databases, while evaluation was done using annotated faces in the wild [73] and FDDB [22] databases.

### 3.2 Database and evaluation metrics

Experiments with the Viola–Jones algorithm have been done in a Samsung 5 Ultra 530U3C A0A laptop (i3-3217U processor, 8 Gb DDR3 RAM, built-in Intel HD Graphics 4000) with MS WIndows 7 Pro. The proposed face tracking algorithm has been implemented in C++ using OpenCV 2.4.8.0 with activated parallelisation. The Viola–Jones algorithm (with MB-LBP features) and the Farnebäck dense optical flow method are from OpenCV as well. The utilised implementation of Viola–Jones uses a cascade of 20 classifiers. Experiments with MTCNN and VGG face detectors have been done in a Dell E7240 laptop (i7-4600 processor, 16 Gb DDR3 RAM, built-in Intel HD Graphics 4400) with MSWindows 8.1 Pro. MTCNN is implemented in Caffe, while VGG is implemented in MatConvNet, and both are called from Matlab r2014b x64. Regarding the experimental dataset, we have used the Boston Head Tracking Database [74], having 45 videos of resolution 320 × 240 recorded at 30 fps under uniform light conditions. Each video has 199 frames and contains one person moving the head around. Ground-truth information is available through the UVAEYES annotations which have the positions of the eyes in each frame (see Fig. 2 for an example) [75]. This database is also employed in some previous face tracking studies, e.g. [32, 76, 77].

The developed face tracker has some parameters that have been set empirically. Viola–Jones is called every $n = 20$ frames and when this is done, the updated likelihood map is computed using $\alpha = 0.5$. The reject level threshold for accepting windows to contribute to the likelihood map has been set to $\tau = 15$, the shrinking factor used is $s = 1/3$, and the binarisation threshold is $c = 65$. The latter value of $c$ is set by knowing that a window containing a face increments corresponding pixels by at least 15 (due to $\tau$), and that Viola–Jones usually output several windows of different sizes centred around the same face. We compare our method with the original Viola–Jones face detector, and with MTCNN and VGG detectors. Accuracy of the algorithms is measured by the Euclidean distance between the computed face centre, and the ground-truth face centre $[x, y]$, which is extracted via following equations:

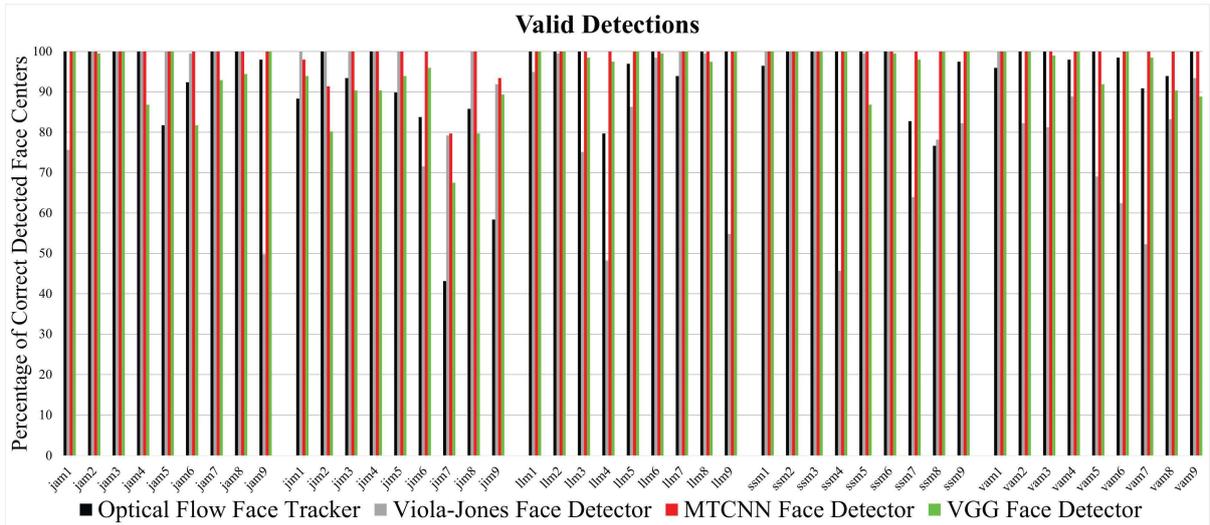

**Fig. 3** *Percentage of correctly detected face centres per video. Better seen in colour*

$$x = x_1 + \frac{x_2 - x_1}{2} \quad (2)$$

$$y = y_1 + \frac{y_2 - y_1}{2} + 10 \quad (3)$$

These calculate the middle point between the eyes from the ground-truth information ($x1$, $y1$ and $x2$, $y2$, respectively), and add 10 pixels in $y$ (vertical) direction to estimate the face centre. Origin of coordinates is in the upper left corner of the image. This works quite well as there is not much difference in the sizes of faces in our database.

The evaluated algorithms output a bounding box with the detected face in a frame (see Fig. 2). The centre [$x_t$, $y_t$] of the bounding box is used as the detected face centre, and it is counted as a valid detection if its distance to the ground truth is lower than 20 pixels, otherwise it is counted as a false positive. In case of multiple detections in a frame, all the distances to the ground truth are computed, and the nearest one is chosen as valid, as long as it is within the threshold defined above. All other detections are counted as false positives. If no bounding boxes are detected in a frame, it is counted as a false negative. It should be noted that only one false negative per frame can occur, while the number of false positives per frame can be higher than one.

### 3.3 Detection rate

The detection rate $r$ of a video is calculated by summing up the valid detections $d_i$ at each frame, and dividing by the number of frames $n_{\text{frames}}$ (ideally $d_i = 1$ in each frame)

$$r = \frac{1}{n_{\text{frames}}} \times \sum_{i=1}^{n_{\text{frames}}} d_i \quad (4)$$

Fig. 3 shows the average detection rate per video of the Boston Head Tracking Database. The average detection rate (across all videos) of the proposed optical flow face tracker is of 93.68%, superior to the Viola–Jones detector, which has an average detection rate of 86.81%. The other two detectors have an average detection rate of 99.17% (MTCNN) and 95.16% (VGG). Therefore, the modifications proposed to the Viola–Jones face tracker brings the detection rate just one step behind newer methods based on CNNs or DPM.

### 3.4 Detection accuracy and false detections

Fig. 4*a* shows the average detection accuracy per video of the face detectors. The false negatives and false positives per video are also given in Fig. 5. The average accuracy over the whole database is of 5.99 pixels (our algorithm), 2.56 pixels (Viola–Jones), 5.45 (MTCNN) and 11.07 (VGG). Our detector shows an accuracy comparable to MTCNN, with both algorithms having very few false negatives (face loss). We do not consider, however, these numbers to be representatively different than the accuracy of Viola–Jones, given the fact that we are estimating the face centre from the ground truth by adding a fixed amount of 10 pixels in vertical direction to the middle point between the two eyes, but this offset may be slightly different for each frame. These numbers, however, are computed considering only the frames where a face is detected, so frames with false negatives are not counting towards these values, but Viola–Jones and VGG show false negatives (face loss) in a number of frames (Fig. 5, top).

In addition to a number of false negatives and positives in some videos, it is worth noting the worse accuracy of the VGG algorithm. A further analysis of the output images given by this detector reveals that it tends to overestimate the face region (see Fig. 2), resulting in bounding boxes that are slightly bigger than the face (even in frontal upright faces, as can be seen in the bottom left image). Due to that, the centre of the bounding box will be displaced, resulting in a worse accuracy, or in a false positive if the distance to the ground truth exceeds our threshold.

As mentioned above, the Viola–Jones detector produces a number of false negatives in some of the videos. This is because a window has to pass the 20 classifiers of the cascade to be considered a face, which may not be the case in certain poses. This condition, on the other hand, causes Viola–Jones to show few false positives. The number of false negatives in the optical flow detector is kept low due to: (i) the likelihood map, which considers if a window has passed most of the classifiers of the cascade (even if it does not pass the 20 classifiers), and (ii) the interpolation of the likelihood map via optical flow, which allows to track the face across frames where Viola–Jones may not detect a face. By design, the optical flow face tracker will always detect a face as long as it is correctly initialised with one. On the other hand, if Viola–Jones produces false negatives, the likelihood map is not refreshed with new face information, causing that the tracker becomes inaccurate and produces false positives over time. For this reason, the videos where our tracker has a high number of false positives are videos where Viola–Jones has a high number of false negatives too (Fig. 5). Fig. 4*b* shows the accuracy of the detectors across frames in the video jam1.avi (Fig. 2). Frames with false negatives are marked with a value of −5. When Viola–Jones loses the face, the optical flow detector is able to track its position, although the distance to the ground-truth centre starts to increase. The latter eventually would cause false positives after a period of time. However, when Viola–Jones detects the face again, the optical flow tracker regains a better accuracy. It can also be observed in Fig. 4*b* the similar accuracy of the optical flow and MTCNN detectors, as indicated above.

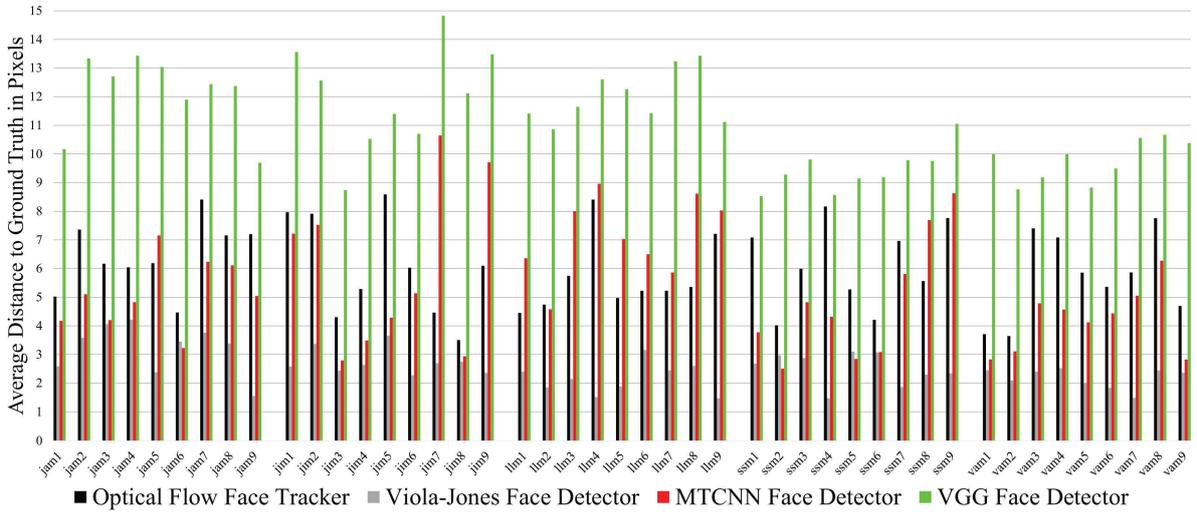

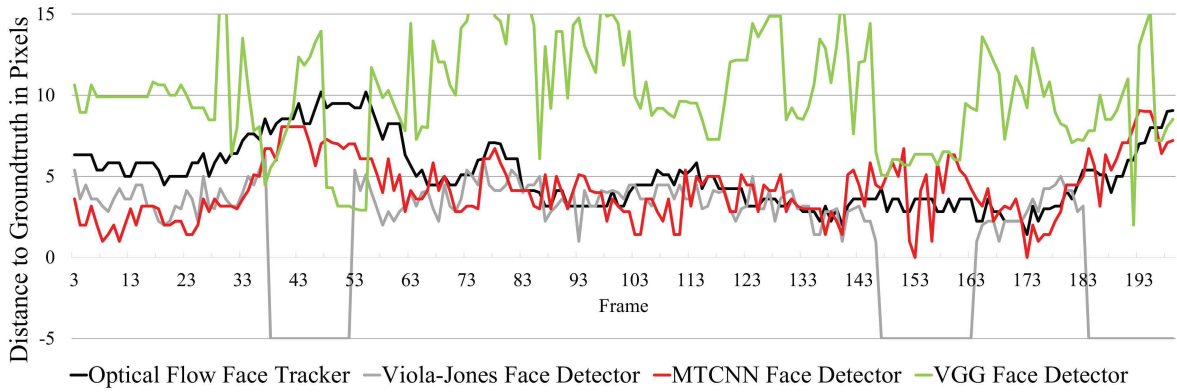

**Fig. 4** *Detection accuracy. Better seen in colour*
*(a)* Average detection accuracy per video, *(b)* Detection accuracy in video jam1.avi

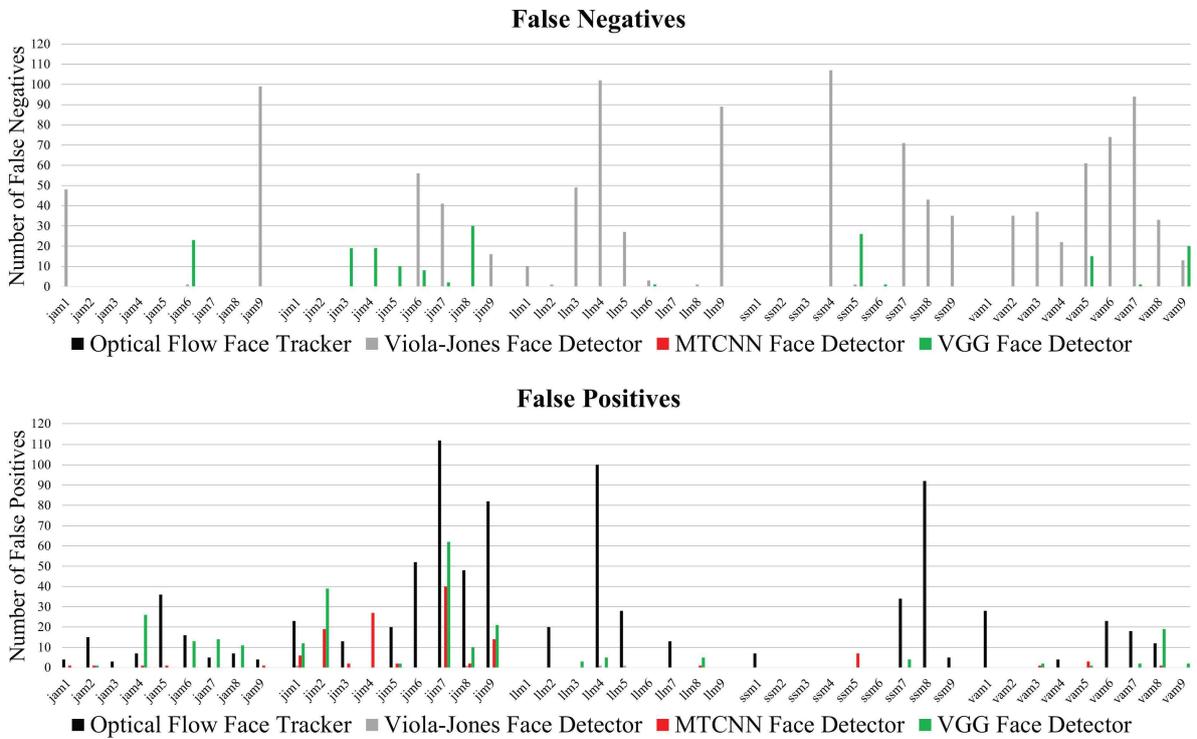

**Fig. 5** *False negatives and positives per video. Better seen in colour*

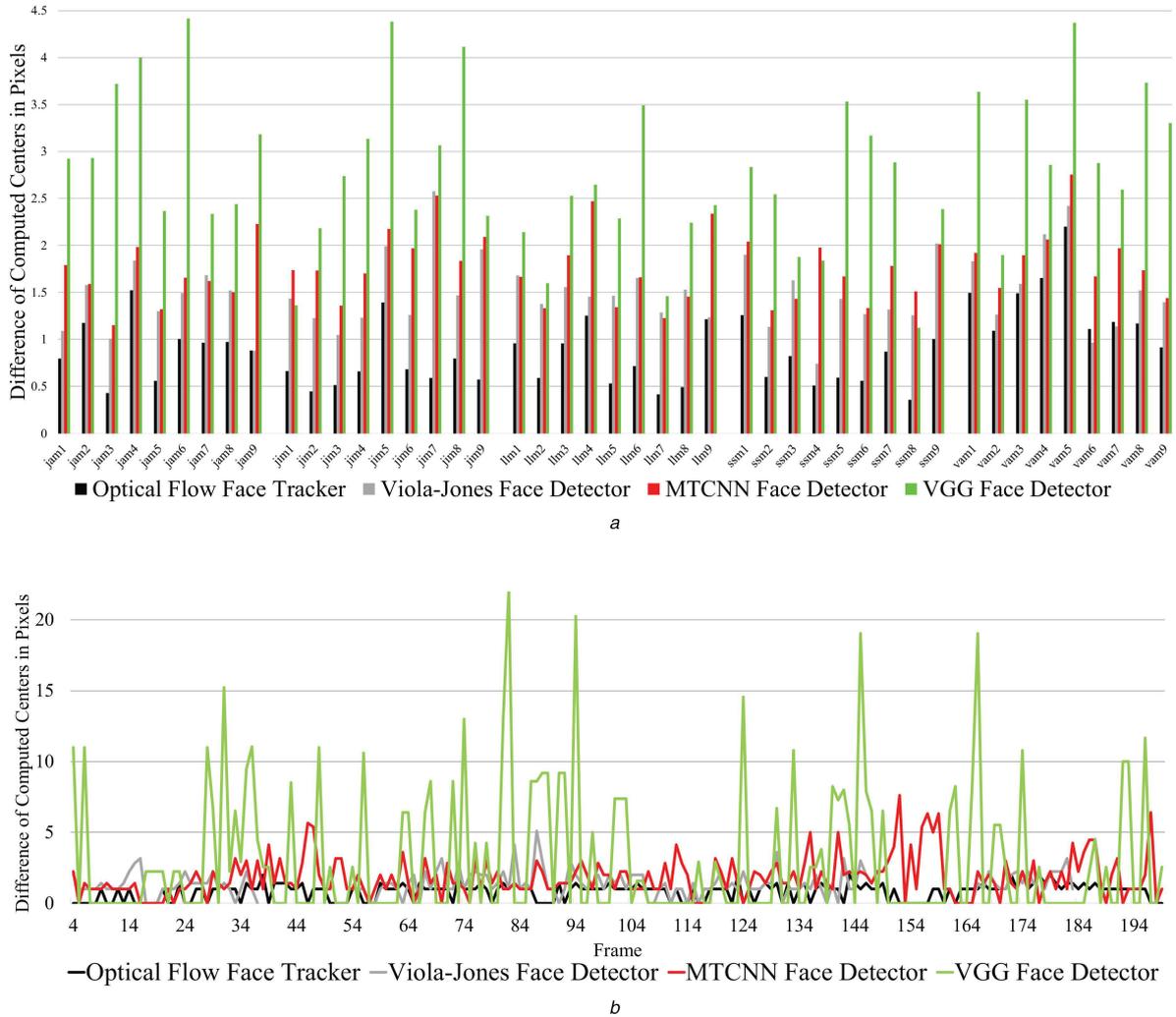

**Fig. 6** *Detection stability. Better seen in colour*
*(a)* Average detection stability per video, *(b)* Detection stability in video jam1.avi

### 3.5 Detection stability

Erratic movements of the face across frames are not desirable in applications, as faces move relatively slow in practice. The stability $e$ is measured as the Euclidean distance between the face centres of two successive frames with valid detections

$$e = \sqrt{(x_t - x_{t-1})^2 + (y_t - y_{t-1})^2} \qquad (5)$$

Fig. 6*a* shows the average stability per video. In general, the optical flow face tracker has less erratic movements. The average on all videos equals 0.9 pixel for the optical flow face tracker, 1.5 pixels for the Viola–Jones detector, 1.8 for MTCNN and 2.8 for VGG. By increasing $\alpha$, which determines the weight of the interpolated likelihood map in (1), the peaks of the optical flow face tracker could be lowered. However, it is likely that the accuracy and the detection rate be worsened. An improvement to further smooth the bounding box coordinates would be to incorporate longer sequences in the computation of the likelihood map via multi-frame optical flow computation [78], although this would imply higher computational cost. Fig. 6*b* shows the stability across frames in the video jam1.avi, further illustrating the higher smoothness of the optical flow tracker in the detection. The values of Viola–Jones algorithm are partly discontinuous due to false negatives in the respective frames.

### 3.6 Detection speed

Speed of the algorithms is measured in milliseconds. The average computation time per frame in each video is shown in Fig. 7*a*. The Viola–Jones algorithm is faster (6 ms in average versus 13 ms), although both are capable of working in real time. The average computation time of the other two algorithms is 137.18 ms (MTCNN) and 253.74 ms (VGG), which is considerably higher despite being executed in a more powerful machine (Section 3.2). The latter two systems could be benefited of GPU acceleration (not used here), although it may not be feasible in some applications involving limited computing resources, e.g. smart-phones. It has to be mentioned too that the average CPU usage of Viola–Jones is higher due to the parallelisation of the OpenCV implementation utilised, resulting in lower computation time. In contrast, the Farnebäck optical flow implementation employed only uses one core, reducing the average CPU usage but increasing the execution time. Fig. 7*b* shows the split between computation of the optical flow (grey) and the rest of our tracking algorithm (blue), where it can be observed that the optical flow takes the majority of the computation time (11.33 ms in average versus 1.67 ms of the remaining operations). Based on the above considerations, however, significant time savings could be achieved by parallelising the optical flow computation, since it is the most time-consuming operation. The proposed algorithm could be also ideally used in applications demanding face detection, and where the optical flow should be computed anyway, as lip-motion analysis [69].

### 3.7 Tracking during occlusions

The developed face tracking algorithm is able to track faces during partial and complete occlusion under certain circumstances, such as fast moving objects passing (like cars). This is because the Farnebäck dense flow technique employed is invariant to very fast motion. Fig. 8 illustrates two examples where the developed face tracker is able to track faces under occlusion. In these two cases, the Viola–Jones algorithm is not able to detect the face.

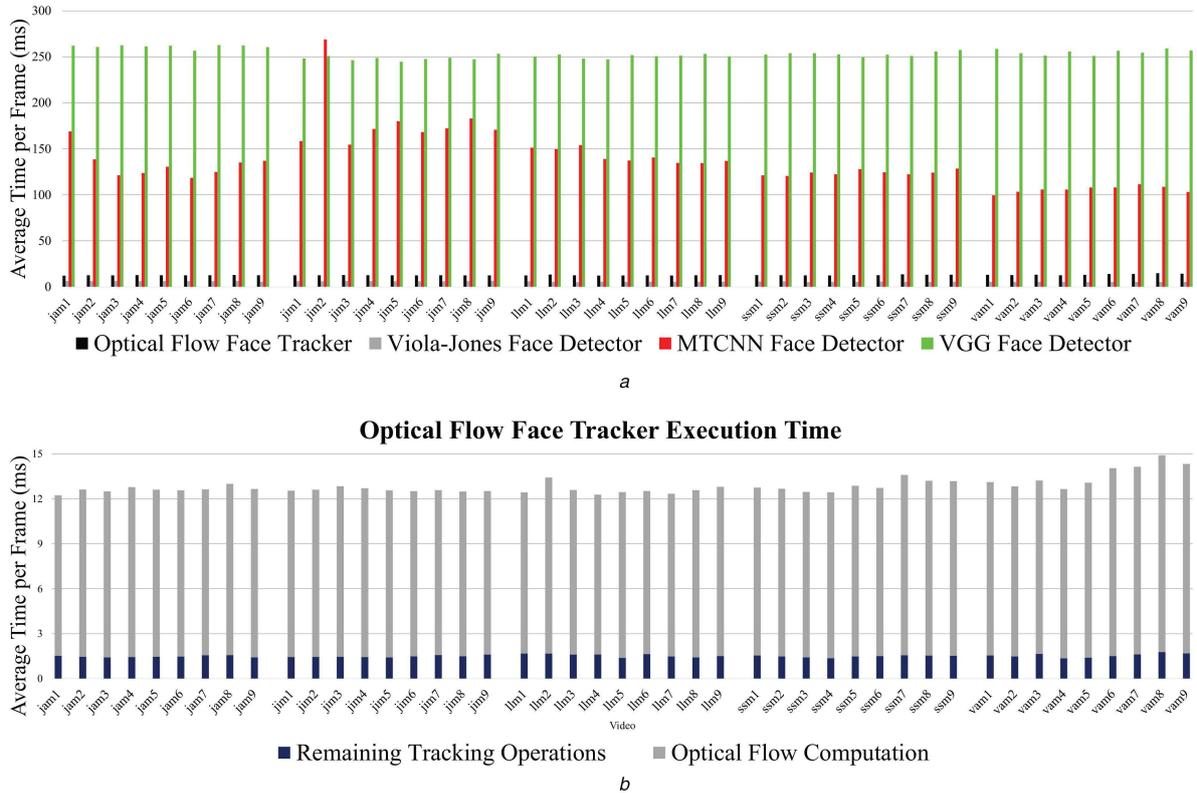

**Fig. 7** *Computation time*
*(a)* Average computation time per frame per video, *(b)* Average computation time per frame per video of the optical flow tracker. Time is split into computation time of optical flow (grey) and of the remaining operations (blue)

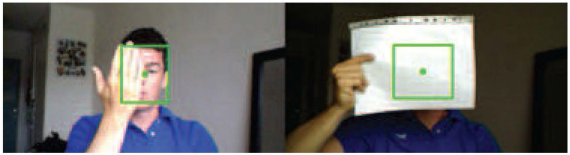

**Fig. 8** *Optical flow face tracker is able to track faces under partial occlusion (left) and also under complete occlusion (right)*

## 4 Conclusion

We present here a novel real-time face tracking algorithm which utilises a modified version of the Viola–Jones algorithm for face detection. In contrast to a pure Viola–Jones face detector, the developed approach calls a modified Viola–Jones method only at every $n = 20$ frames for refreshing a likelihood map. The likelihood values within this map are dependent on the numbers of classification stages which each detection window passes, thus preserving information about near-positives. In order to track faces, the likelihood map is interpolated frame by frame with a flow map computed by the Farnebäck dense optical flow method. The resulting likelihood map of the modified Viola–Jones algorithm contributes to the system's likelihood map by recursive filtering every $n$ frames.

The developed face tracking algorithm and the original Viola–Jones face detector have been evaluated on the Boston Head Tracking Database. We also use in our benchmark two recently published face detection algorithms based on MTCNNs and DPM, the latter included in the release of the VGG-Face CNN recognition descriptor (VGG). The developed face tracker achieves a higher detection rate than the Viola–Jones face detector, and just one step behind the other two methods. The valid detections of our method are less accurate than Viola–Jones, but on average the distance to ground-truth face centre is below 6 pixels. This is a very good value considering that the coordinates for the face centre are derived from ground-truth information of eye coordinates. The accuracy of our method is comparable to MTCNN, while VGG tends to overestimate the size of the face bounding box, resulting in worse accuracy and a higher number of false negatives. Furthermore, the optical flow face tracker shows less erratic movements of detections than any other detector.

In terms of false detections, the Viola–Jones detector shows more false negatives, because it demands that candidate windows have to pass the whole cascade of classifiers to be considered a face, which on the other hand results in less false positives. Our system shows zero false negatives since it allows candidate windows to pass only a number of classifiers of the cascade, and once a face window is detected, it is interpolated to subsequent frames via optical flow. Compared with the original Viola–Jones implementation, the likelihood map approach enables faces to be detected even when they do not pass all of the stages of the cascade classifier. Due to the fact that the likelihood map is never discarded completely, a region gets also a high value in the likelihood map if the respective window passes, for example 17, 18 or 19 stages within several executions of the modified Viola–Jones method, allowing to reduce false negatives. Another advantage of the developed face tracker is that it can also track faces under partial and complete occlusion. On the other hand, our system is dependent of the success of Viola–Jones in finding a face, so in videos where Viola–Jones has many false negatives, our system has a higher number of false positives since it cannot receive a refresh frame.

In comparison to the original Viola–Jones algorithm, our face tracker needs more computation time. The Farnebäck flow computation takes a majority of computation time as this method has been used 'as is' in OpenCV. The demand in computational power is caused by the fact that Farnebäck flow is not parallelised in OpenCV, whereas the Viola–Jones algorithm is. Subtracting the computation time of Farnebäck flow from the face tracker's computation time, our system is about four times faster than the Viola–Jones detector. This fact makes the face tracker an appropriate tool where in applications where dense flow computation has to be done anyway and where face tracking is needed, e.g. lip analysis [69]. By optimising the Farnebäck flow computation method, the tracker execution time could be lowered drastically as well. The MTCNN and VGG detectors, on the other hand, have an execution time one order of magnitude higher than Viola–Jones and our face tracker. These two systems could be benefitted of GPU acceleration, since they are implemented using CNN frameworks, although it may not be feasible in some applications with limited resources.

The developed tracker relies on the Viola–Jones algorithm, which means that errors or inaccuracies of the Viola–Jones algorithm during initialisation or refresh of the likelihood map is propagated to our system. This may result in false positives after a period of time if Viola–Jones is not able to detect a face when it is called every $n = 20$ frames. This effect could be minimised for example by triggering other methods for face detection with higher accuracy if Viola–Jones fails, or by using a detection algorithm with lower computation time in each frame to quick check if a face is lost when interpolating the likelihood map and if so, call Viola–Jones earlier than every 20 frames. Another possibility is to execute a modified version of Viola–Jones only on the extracted face areas, in order to check if the tracker has lost the faces or not. A complementary strategy to improve the accuracy of our method would be to make the weighting parameter $\alpha$ in (1) dependent on the detection confidence of the algorithm used in a given frame, so the likelihood map is refreshed selectively in proportion to the confidence on the detection. A good algorithm with good detection accuracy and few false negatives could be the MTCNN method employed in this paper, although it should be considered that its execution time is considerably higher than Viola–Jones. To further improve the accuracy and stability of the detection, we are also exploring the use of multi-frame optical flow computation in order to consider longer sequences in the computation of the likelihood map [78].

The Viola–Jones framework is not perfectly suited for the implementation of our interpolated likelihood map, as some extra amount of tweaking was necessary to build up an initial map from the detection results. Viola–Jones was chosen for this work as it is the most well known and widely used approach to static face detection today, and it is able to work in real time [26]. However, this is rapidly changing as deep learning approaches are steadily improving on performance, with hardware dedicated to them being developed jointly (embedded GPUs). Although Viola–Jones works very well with near-frontal faces, its performance significantly suffers with faces having arbitrary poses or other degradations. It could be expected that accuracy of MTCNN, VGG or other approaches for unconstrained face detection would be improved in the same way that Viola–Jones has been improved here. It is an interesting future approach to replace both Farnebäck flow and Viola Jones with deep learning methods, such as the MTCNN approach evaluated here [7], although some tweaking would be also necessary. DeepFace [79] is very promising as it improves performance greatly, it is even better adapted to detecting identity (not just face occurrence) and it outputs a likelihood map to use. Together with recent advances in optical flow computation using deep learning approach, especially FlowNet [80], we have a promising framework that is much better adapted to our likelihood map generation.

## 5 Acknowledgments

Author Fernando Alonso-Fernandez thanks the Swedish Research Council for funding his research. Authors acknowledge the CAISR program and the SIDUS-AIR project of the Swedish Knowledge Foundation.

## 6 References

[1] Li, S.Z., Jain, A.K. (Eds.): 'Handbook of face recognition' (Springer Verlag, 2011, 2nd edn.)
[2] Jain, A.K., Kumar, A.: 'Biometrics of next generation: an overview', in Mordini, E., Tzovaras, D. (Eds.): 'Second generation biometrics' (Springer, 2010)
[3] Stallkamp, J., Ekenel, H.K., Stiefelhagen, R.: 'Video-based face recognition on real-world data'. Proc. Int. Conf. Computer Vision, ICCV, October 2007, pp. 1–8
[4] Viola, P., Jones, M.: 'Rapid object detection using a boosted cascade of simple features'. Proc. Computer Vision and Pattern Recognition Conf., CVPR, December 2001, vol. 1, pp. 511–518
[5] Li, S.Z.: 'Face detection', in Li, S.Z., Jain, A. (Eds.): 'Handbook of face recognition' (Springer Verlag, 2011, 2nd edn.)
[6] Ranftl, A., Alonso-Fernandez, F., Karlsson, S.: 'Face tracking using optical flow'. Proc. Int. Conf. the Biometrics Special Interest Group, BIOSIG (Best Paper Award), September 2015, pp. 1–5
[7] Zhang, K., Zhang, Z., Li, Z., et al.: 'Joint face detection and alignment using multitask cascaded convolutional networks', IEEE Signal Process. Lett., 2016, 23, (10), pp. 1499–1503
[8] Parkhi, O.M., Vedaldi, A., Zisserman, A.: 'Deep face recognition'. Proc. British Machine Vision Conf., BMVC, 2015
[9] Mathias, M., Benenson, R., Pedersoli, M., et al.: 'Face detection without bells and whistles'. Proc. European Conf. Computer Vision, ECCV, 2014, pp. 720–735
[10] Yang, M.-H., Kriegman, D., Ahuja, N.: 'Detecting faces in images: a survey', IEEE Trans. Pattern Anal. Mach. Intell., 2002, 24, (1), pp. 34–58
[11] Rowley, H.A., Baluja, S., Kanade, T.: 'Neural network-based face detection', IEEE Trans. Pattern Anal. Mach. Intell., 1998, 20, (1), pp. 23–38
[12] Sung, K.-K., Poggio, T.: 'Example-based learning for view-based human face detection', IEEE Trans. Pattern Anal. Mach. Intell., 1998, 20, (1), pp. 39–51
[13] Li, Y.M., Gong, S.G., Liddell, H.: 'Support vector regression and classification based multi-view face detection and recognition'. Proc. Int. Conf. Automatic Face and Gesture Recognition, FG, 2000
[14] Viola, P., Jones, M.: 'Robust real-time face detection', Int. J. Comput. Vis., 2004, 57, (2), pp. 137–154
[15] Teferi, D., Bigun, J.: 'Evaluation protocol for the dxm2vts database and performance comparison of face detection and face tracking on video'. Int. Conf. Pattern Recognition, ICPR, December 2008, pp. 1–4
[16] Lienhart, R., Kuranov, A., Pisarevsky, V.: 'Empirical analysis of detection cascades of boosted classifiers for rapid object detection'. Proc. DAGM 25th Pattern Recognition Symp., 2003
[17] Zhang, L., Chu, R., Xiang, S., et al.: 'Face detection based on multi-block lbp representation'. Proc. Int. Conf. Biometrics, ICB, 2007
[18] Li, H., Lin, Z., Shen, X., et al.: 'A convolutional neural network cascade for face detection'. Proc. Int. Conf. Computer Vision and Pattern Recognition, CVPR, June 2015, pp. 5325–5334
[19] Liu, Z., Luo, P., Wang, X., et al.: 'Deep learning face attributes in the wild'. Proc. Int. Conf. Computer Vision, ICCV, 2015, pp. 3730–3738
[20] Yang, S., Luo, P., Loy, C.C., et al.: 'From facial parts responses to face detection: a deep learning approach'. Proc. Int. Conf. Computer Vision, ICCV, 2015, pp. 3676–3684
[21] Zhang, C., Zhang, Z.: 'Improving multiview face detection with multi-task deep convolutional neural networks'. Proc. IEEE Winter Conf. Applications of Computer Vision, March 2014, pp. 1036–1041
[22] Jain, V., Learned-Miller, E.G.: 'FDDB: a benchmark for face detection in unconstrained settings'. Technical Report UMCS-2010-009, University of Massachusetts, Amherst, 2010
[23] Köstinger, M., Wohlhart, P., Roth, P.M., et al.: 'Annotated facial landmarks in the wild: a large-scale, real-world database for facial landmark localization'. Proc. Int. Conf. Computer Vision Workshops, ICCV, November 2011, pp. 2144–2151
[24] Li, J., Wang, T., Zhang, Y.: 'Face detection using surf cascade'. Proc. Int. Conf. Computer Vision Workshops, ICCV, November 2011, pp. 2183–2190
[25] Li, H., Hua, G., Lin, Z., et al.: 'Probabilistic elastic part model for unsupervised face detector adaptation'. Proc. Int. Conf. Computer Vision, ICCV, December 2013, pp. 793–800
[26] Liao, S., Jain, A.K., Li, S.Z.: 'A fast and accurate unconstrained face detector', IEEE Trans. Pattern Anal. Mach. Intell., 2016, 38, (2), pp. 211–223
[27] Zhao, W., Chellappa, R., Phillips, P., et al.: 'Face recognition: a literature survey', ACM Comput. Surv., 2003, 35, (4), pp. 399–458
[28] Dornaika, F., Ahlberg, J.: 'Fast and reliable active appearance model search for 3-d face tracking', IEEE Trans. Syst. Man Cybern. B, Cybern., 2004, 34, (4), pp. 1838–1853
[29] Dornaika, F., Ahlberg, J.: 'Fitting 3d face models for tracking and active appearance model training', Image Vis. Comput., 2006, 24, (9), pp. 1010–1024
[30] Maurel, P., McGonigal, A., Keriven, R., et al.: '3d model fitting for facial expression analysis under uncontrolled imaging conditions'. Proc. Int. Conf. Pattern Recognition, ICPR, December 2008, pp. 1–4
[31] Roy-Chowdhury, A.-K., Xu, Y.: 'Face tracking', in Li, S.Z., Jain, A. (Eds.): 'Encyclopedia of biometrics' (Springer, 2015, 2nd edn.)
[32] Gast, E.R.: 'A framework for real-time face and facial feature tracking using optical flow pre-estimation and template tracking'. MS thesis, Leiden University, 2010
[33] DeCarlo, D., Metaxas, D.: 'Optical flow constraints on deformable models with applications to face tracking', Int. J. Comput. Vis., 2000, 38, (2), pp. 99–127
[34] Delmas, P., Eveno, N., Lievin, M.: 'Towards robust lip tracking'. Proc. Int. Conf. Pattern Recognition, ICPR, 2002, vol. 2, pp. 528–531
[35] Wu, Z., Aleksic, P.S., Katsaggelos, A.K.: 'Lip tracking for mpeg-4 facial animation'. Proc. IEEE Int. Conf. Multimodal Interfaces, 2002, pp. 293–298
[36] Chen, J., Tiddeman, B.: 'Multi-cue facial feature detection and tracking'. Proc. Int. Conf. Image and Signal Processing, ICISP, 2008, pp. 356–367
[37] Kalal, Z., Mikolajczyk, K., Matas, J.: 'Face-TLD: tracking-learning-detection applied to faces'. Proc. Int. Conf. Image Processing, ICIP, 2010, pp. 3789–3792
[38] Comaniciu, D., Ramesh, V., Meer, P.: 'Kernel-based object tracking', IEEE Trans. Pattern Anal. Mach. Intell., 2003, 25, (5), pp. 564–577
[39] Tao, J., Tan, Y.-P.: 'A unified probabilistic approach to face detection and tracking'. IEEE Int. Symp. Circuits and Systems, SCAS, May 2005, vol. 4, pp. 3797–3800

[40] Huang, H.-P., Lin, C.-T.: 'Multi-camshift for multi-view faces tracking and recognition'. Proc. IEEE Int. Conf. Robotics and Biomimetics, ROBIO, December 2006, pp. 1334–1339

[41] Nguyen, Q.-A., Robles-Kelly, A., Shen, C.: 'Enhanced kernel-based tracking for monochromatic and thermographic video'. IEEE Int. Conf. Video and Signal Based Surveillance, AVSS, November 2006, pp. 28–28

[42] Vadakkepat, P., Lim, P., De-Silva, L.C., *et al.*: 'Multimodal approach to human-face detection and tracking', *IEEE Trans. Ind. Electron.*, 2008, **55**, (3), pp. 1385–1393

[43] Hiremath, P.S., Hiremath, M., Mahesh, R.: 'Face detection and tracking in video sequence using fuzzy geometric face model and mean shift', *Int. J. Adv. Trends Comput. Sci. Eng.*, 2013, **2**, (1), pp. 41–46

[44] Grest, D., Koch, R.: 'Realtime multi-camera person tracking for immersive environments'. Proc. IEEE Workshop on Multimedia Signal Processing, September 2004, pp. 387–390

[45] Su, C., Huang, L.: 'Spatio-temporal graphical-model-based multiple facial feature tracking', *EURASIP J. Appl. Signal Process.*, 2005, **2005**, pp. 2091–2100

[46] Yun, T., Guan, L.: 'Fiducial point tracking for facial expression using multiple particle filters with kernel correlation analysis'. Proc. IEEE Int. Conf. Image Processing, ICIP, September 2010, pp. 373–376

[47] Chae, Y.N., Ha, J., Yang, H.S.: 'Development of an efficient face detection and tracking system for mobile devices'. Proc. IEEE Int. Conf. Virtual Systems and Multimedia, VSMM, October 2010, pp. 192–196

[48] Zhu, Z., Ji, Q.: '3d face pose tracking from an uncalibrated monocular camera'. Proc. Int. Conf. Pattern Recognition, ICPR, August 2004, vol. **4**, pp. 400–403

[49] Girondel, V., Caplier, A., Bonnaud, L.: 'Real time tracking of multiple persons by Kalman filtering and face pursuit for multimedia applications'. Proc. IEEE Southwest Symp. Image Analysis and Interpretation, March 2004, pp. 201–205

[50] Destrero, A., Odone, F., Verri, A.: 'A system for face detection and tracking in unconstrained environments'. Proc. IEEE Conf. Advanced Video and Signal Based Surveillance, AVSS, September 2007, pp. 499–504

[51] Bing, X., Wei, Y., Chareonsak, C.: 'Automatic focusing technique for face detection and face contour tracking'. IEEE Int. Workshop on Biomedical Circuits and Systems, December 2004, pp. S3/2-9–S3/2-12

[52] Mohabbati, B., Kasaei, S.: 'Face localization and versatile tracking in wavelet domain'. Proc. IEEE Conf. Information and Communication Technologies, ICTTA, 2006, vol. **1**, pp. 1552–1556

[53] Loutas, E., Nikou, C., Pitas, I.: 'An information theoretic approach to joint probabilistic face detection and tracking'. Proc. Int. Conf. Image Processing, ICIP, 2002, vol. **1**, pp. I–505–I–508

[54] Wu, B., Hu, B.-G., Ji, Q.: 'A coupled hidden markov random field model for simultaneous face clustering and tracking in videos', *Pattern Recognit.*, 2017, **64**, pp. 361–373

[55] Subban, R., Muthukumar, S., Pasupathi, P., *et al.*: 'Face tracking techniques in color images: a study and review', *Int. J. Eng. Res. Technol.*, 2013, **2**, (12), pp. 2481–2487

[56] Jagathishwaran, R., Ravichandran, K.S., Jayaraman, P.: 'A survey on face detection and tracking', *World Appl. Sci. J.*, 2014, **29**, pp. 140–145

[57] Kristan, M., Pflugfelder, R., Leonardis, A., *et al.*: '*The visual object tracking VOT2014 challenge results*' (Springer International Publishing, Cham, 2015), pp. 191–217

[58] Wu, Y., Lim, J., Yang, M.H.: 'Object tracking benchmark', *IEEE Trans. Pattern Anal. Mach. Intell.*, 2015, **37**, (9), pp. 1834–1848

[59] My, V.D., Zell, A.: 'Real time face tracking and pose estimation using an adaptive correlation filter for human-robot interaction'. Proc. European Conf. Mobile Robots, September 2013, pp. 119–124

[60] Zhou, L.B., Wang, H., Mou, W., *et al.*: 'Robust face detection and tracking under natural conditions'. Proc. Int. Conf. Robotics and Biomimetics, ROBIO, December 2013, pp. 934–939

[61] Gaxiola, L.N., Díaz-Ramírez, V.H., Tapia, J.J., *et al.*: '*Robust face tracking with locally-adaptive correlation filtering*' (Springer International Publishing, Cham, 2014), pp. 925–932

[62] Galoogahi, H.K., Sim, T., Lucey, S.: 'Correlation filters with limited boundaries'. Proc. Conf. Computer Vision and Pattern Recognition, CVPR, June 2015, pp. 4630–4638

[63] Chen, W., Zhang, K., Liu, Q.: 'Robust visual tracking via patch based kernel correlation filters with adaptive multiple feature ensemble', *Neurocomputing*, 2016, **214**, pp. 607–617

[64] Bigun, J.: '*Vision with direction*' (Springer, 2006)

[65] Shi, J., Tomasi, C.: 'Good features to track'. Proc. Int. Conf. Computer Vision and Pattern Recognition, CVPR, 1994

[66] Bigun, J., Granlund, G.H.: 'Optimal orientation detection of linear symmetry'. Proc. Int. Conf. Computer Vision, ICCV, June 1987, pp. 433–438

[67] Harris, C., Stephens, M.: 'A combined corner and edge detector'. Proc. Fourth Alvey Vision Conf., 1988, pp. 147–151

[68] Lucas, B.D., Kanade, T.: 'An iterative image registration technique with an application to stereo vision'. Proc. Int. Joint Conf. Artificial Intelligence, IJCAI, October 1981, pp. 674–679

[69] Karlsson, S.M., Bigun, J.: 'Lip-motion events analysis and lip segmentation using optical flow'. Proc. IEEE Computer Vision and Pattern Recognition Biometrics Workshop, CVPRW, 2012

[70] Farneback, G.: 'Two-frame motion estimation based on polynomial expansion'. Proc. Swedish Symp. Image Analysis, SSBA, 2003

[71] Yang, S., Luo, P., Loy, C.C., *et al.*: 'Wider face: a face detection benchmark'. Proc. Conf. Computer Vision and Pattern Recognition, CVPR, 2016

[72] Yan, J., Zhang, X., Lei, Z., *et al.*: 'Face detection by structural models', *Image Vis. Comput.*, 2014, **32**, (10), pp. 790–799, Best of Automatic Face and Gesture Recognition 2013

[73] Zhu, X., Ramanan, D.: 'Face detection, pose estimation, and landmark localization in the wild'. Proc. Conf. Computer Vision and Pattern Recognition, CVPR, June 2012, pp. 2879–2886

[74] La Cascia, M., Sclaroff, S., Athitsos, V.: 'Fast, reliable head tracking under varying illumination: an approach based on registration of texture-mapped 3d models', *IEEE Trans. Pattern Anal. Mach. Intell.*, 2000, **22**, (4), pp. 322–336

[75] Valenti, R., Gevers, T.: 'Robustifying eye center localization by head pose cues'. Proc. Int. Conf. Computer Vision and Pattern Recognition, CVPR, 2009

[76] Jung, S.U., Nixon, M.S.: 'Model-based feature refinement by ellipsoidal face tracking'. Proc. Int. Conf. Pattern Recognition, ICPR, November 2012, pp. 1209–1212

[77] Duong, C.N., Dinh, T.C.P., Ngo, T.D., *et al.*: 'Robust eye localization in video by combining eye detector and eye tracker'. Proc. Int. Conf. Pattern Recognition, ICPR, November 2012, pp. 242–245

[78] Irani, M.: 'Multi-frame correspondence estimation using subspace constraints', *Int. J. Comput. Vis.*, 2002, **48**, (3), pp. 173–194

[79] Taigman, Y., Yang, M., Ranzato, M., *et al.*: 'Deepface: closing the gap to human-level performance in face verification'. Proc. Conf. Computer Vision and Pattern Recognition, CVPR, June 2014, pp. 1701–1708

[80] Fischer, P., Dosovitskiy, A., Ilg, E., *et al.*: 'Flownet: learning optical flow with convolutional networks'. Int. Conf. Computer Vision, ICCV, 2015